\documentclass[runningheads,a4paper]{llncs}

\usepackage{amssymb}
\usepackage{graphicx}
\usepackage{subfigure}
\usepackage{caption}
\makeatletter
\renewcommand{\@thesubfigure}{\hskip\subfiglabelskip}
\makeatother

\usepackage{url}
\urldef{\mailsa}\path|{alfred.hofmann, ursula.barth, ingrid.haas, frank.holzwarth,|
\urldef{\mailsb}\path|anna.kramer, leonie.kunz, christine.reiss, nicole.sator,|
\urldef{\mailsc}\path|erika.siebert-cole, peter.strasser, lncs}@springer.com|

\usepackage{indentfirst}
\usepackage{amsmath}
\DeclareMathOperator{\sign}{sign}
\begin{document}

\mainmatter  

\title{CAAD 2018: Iterative Ensemble Adversarial Attack}


%
%
\author{Jiayang Liu\thanks{5th Place in Targeted Adversarial Attack Competition: Team RNG, Team members : Jiayang Liu} \and Weiming Zhang \and Nenghai Yu}


\institute{School of Information Science and Technology, University of Science and Technology of China}

%
%

\maketitle

\begin{abstract}
Deep Neural Networks (DNNs) have recently led to significant improvements in many fields. However, DNNs are vulnerable to adversarial examples which are samples with imperceptible perturbations while dramatically misleading the DNNs. Adversarial attacks can be used to evaluate the robustness of deep learning models before they are deployed. Unfortunately, most of existing adversarial attacks can only fool a black-box model with a low success rate. To improve the success rates for black-box adversarial attacks, we proposed an iterated adversarial attack against an ensemble of image classifiers. With this method, we won the 5th place in CAAD 2018 Targeted Adversarial Attack competition.
\end{abstract}

\section{Introduction}

Deep Neural Networks (DNNs) have recently led to significant improvements in many fields, such as image classification \cite{russakovsky2015imagenet,he2016deep} and speech recognition \cite{amodei2016deep}. However, the generalization properties of the DNNs have been recently questioned because these machine learning models are vulnerable to adversarial examples \cite{szegedy2014intriguing}. An adversarial example is a slightly modified sample that is intended to cause an error output of the DNN based model. In the context of classification task, the adversarial example is crafted to force a model to classify it into a class different from the legitimate class. In addition, adversarial examples have cross-model transferability property \cite{Goodfellow2015Explaining}. The adversarial examples crafted for one model remain adversarial for others, thus making black-box attacks practical in real-world applications. Unfortunately, most of existing adversarial attacks can only fool a black-box model with a low success rate. To improve the success rates for black-box adversarial attacks, we proposed an iterated adversarial attack against an ensemble of image classifiers.

Adversarial attacks are divided into two types: targeted attack and untargeted attack. In targeted attack, the attacker generates adversarial examples which are misclassified by the classifier into a particular class. In untargeted attack, the attacker generates adversarial examples which are misclassified by the classifier into any class as long as it is different from the true class.

\section{Related Work}

\subsection{Fast Gradient Sign Method}
Goodfellow et al. \cite{Goodfellow2015Explaining} propose the Fast Gradient Sign Method (FGSM) for generating adversarial examples. This method uses the derivative of the loss function of the model pertaining to the input feature vector. Given the input image $X$, FGSM is to perturb the gradient direction of each feature by the gradient. Then the classification result of the input image will be changed. For a neural network with cross-entropy cost function $J(X,y)$ where $X$ is the input image and $y_{t}$ is the target class for the input image, the adversarial example is generated as\begin{equation}X^{adv}=X-\epsilon{\sign(\nabla_XJ(X,y_{t}))}\end{equation}where $\epsilon$ is a parameter to determine the perturbation size.

\subsection{Iterative Gradient Sign Method}
The Iterative Gradient Sign Method (IGSM) is the iterative version of FGSM. This method applies FGSM many times with small perturbation size instead of applying adversarial noise with one large perturbation size. The adversarial example of the iterative gradient sign method is generated as\begin{equation}\begin{split}X_0^{adv}&=X,\\X_{N+1}^{adv}&=Clip_{X,\epsilon}\{X_N^{adv}-\alpha{\sign(\nabla_XJ(X_N^{adv},y_{t}))}\}\end{split}\end{equation}where $Clip_{X,\epsilon}\{X'\}$ represents a clipping of the values of the adversarial example. So the results are within $\epsilon$-neighbourhood of the input image $X$. This attack is more powerful because the attacker can control how far the adversarial example past the classification boundary. It was demonstrated that the attack of IGSM was better than FGSM on ImageNet dataset \cite{kurakin2016adversarial}.


\subsection{Carlini \& Wagner Method}
This method is named after its authors \cite{Carlini2017Towards}. The attack can be targeted or untargeted, and has three metrics to measure its distortion (${l_0}$ norm, ${l_2}$ norm and ${l_\infty }$ norm). The authors point out that the untargeted ${l_2}$ norm version has the best performance. It generates adversarial examples by solving the following optimization problem:\begin{equation}\begin{split}\mathop {{\rm{minimize}}}\limits_\delta  {\rm{  }}{\left\| \delta  \right\|_2} &+ c \cdot f(x + \delta )\\s.t. \quad {\rm{  }}x +& \delta  \in {\left[ {0,1} \right]^n}\end{split}\end{equation}

This attack is to look for the smallest perturbation measured by ${l_2}$ norm and make the network classify the image incorrectly at the same time. $c$ is a hyperparameter to balance the two parts of equation $\left( 3 \right)$. The best way to choose $c$ is to use the smallest value of $c$ for which the resulting solution $x + \delta$ has $f\left( {x + \delta } \right) \le 0$. $f(x)$ is the loss function to measure the distance between the input image and the adversarial image. $f(x)$ is defined as: \begin{equation}f(x) = \max (Z{(x)_{true}} - {\max _{i \ne true}}\{ Z{(x)_i}\} , - \kappa )\end{equation}

$Z(x)$ is the pre-softmax classification result vector. $\kappa$ is a hyper-parameter called confidence. Higher confidence encourages the attack to search for adversarial examples that are stronger in classification confidence. High-confidence attacks often have larger perturbations and better transferability to other models.

\section{Proposed Method}

\subsection{Ensemble Attack Method}
An intriguing property of adversarial examples is that adversarial examples generated for one classifier transfer to other classifiers \cite{Goodfellow2015Explaining}. Therefore, a natural method for effective adversarial attacks against unknown classifiers is to generate strong adversarial examples against a large collection of classifiers. We propose to perform an iterated targeted adversarial attack against an ensemble of image classifiers. In our solution we give equal weights to each of classifiers. The classifiers in the ensemble are: Inception V3 \cite{szegedy2016rethinking}, ResNet 50 \cite{he2016deep}, ResNet 101 \cite{he2016deep}, Inception ResNet V2 \cite{szegedy2017inception}, ensemble adversarially trained Inception ResNet V2 (EnsAdv Inception ResNet V2) \cite{tramer2017ensemble} and adversarially trained Inception V3 (Adv Inception V3) \cite{kurakin2016machine}. Our method builds iterative attacks on a loss ensemble neural networks focusing on the classifiers that are difficult to perturb.

It is a common way to combine a number of classifiers by probability ensemble. However, it is usually hard to optimize the loss function because the parameters of different classifiers are coupled inside the logarithm. \begin{equation}{J_{prob}}\left( {X,y} \right) =  - \sum\limits_i^N {{y_j}\log \left( {\frac{1}{M}\sum\limits_i^M {{p_{ij}}\left( X \right)} } \right)}\end{equation}

An upper bound is obtained for the loss function by Jensen's inequality. We propose to optimize the upper bound instead of minimizing ${J_{prob}}\left( {X,y} \right)$. This way of combining classifiers in Eq.\ref{1} is called loss ensemble. The parameters of different neural networks are decoupled in the new loss function ${J_{loss}}\left( {X,y} \right)$, which is benefit to the optimization.
\begin{equation}{J_{prob}}\left( {X,y} \right) \le  - \frac{1}{M}\sum\limits_j^N {\sum\limits_i^M {{y_{ij}}\log \left( {{p_{ij}}\left( X \right)} \right)} }  = {J_{loss}}\left( {X,y} \right)\label{1}\end{equation}

\subsection{Iterative Ensemble Attack}
\begin{figure}[htbp]
  \centering
  \includegraphics[width=1.0\textwidth]{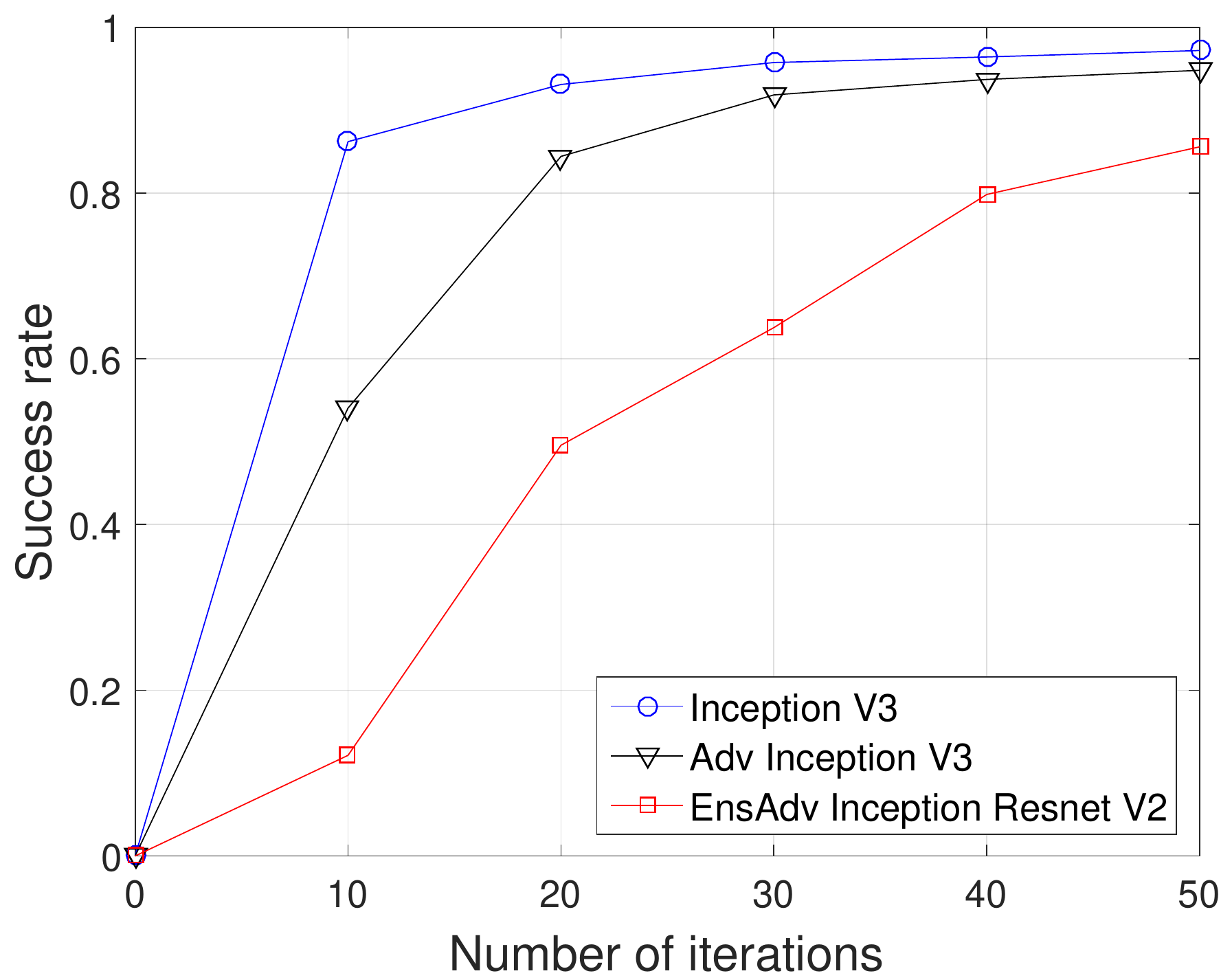}
  \caption{Iterative ensemble attack results for three selected models.}
  \label{iterations}
\end{figure}

The proposed iterative ensemble attack is based on IGSM. We compare our iterative attack method with different number of iterations. Figure \ref{iterations} demonstrates that attacking an adversarial trained model at high success rate requires more iterations than a normal model. For tasks where computation is limited, we preassign the number of iterations for each model or dynamically adjust whether a model is included in each step of the attack by observing if the loss function for that model is small enough. In Eq.\ref{2}, ${\delta _{Nk}} \in \{ 0,1\}$ determines whether to include the loss for model $k$ in the total loss at step $N$.

\begin{equation}\begin{split}X_0^{adv}&=X,\\X_{N + 1}^{adv} &= Clip_{X,\epsilon}\left\{ {X_N^{adv} - \alpha \sign\left( {\frac{1}{M}\sum\limits_k^M {{\delta _{Nk}}{\nabla _X}{J_k}\left( {X_N^{adv},{y_t}} \right)} } \right)} \right\}\end{split}\label{2}\end{equation}

\section{Conclusion}
In this paper, we propose an iterated adversarial attack against an ensemble of image classifiers, which can improve the success rates for black-box adversarial attacks. With this method, we won the 5th place in CAAD 2018 Targeted Adversarial Attack competition. It is easy to make our attack stronger by adding more models or increasing the number of iterations.


\end{document}